\documentclass{article}

\usepackage[preprint]{corl_2021} 
\usepackage{graphicx}
\usepackage{makecell, booktabs}
\usepackage{multirow}
\usepackage{algorithm}
\usepackage[noend]{algpseudocode}
\usepackage{amsmath}
\usepackage{amsfonts,amssymb}

\title{Generalization of Reinforcement Learning with Policy-Aware Adversarial Data Augmentation}

\author{
	Hanping Zhang, \quad  Yuhong Guo\\
	Carleton University, Ottawa, Canada\\
	{jagzhang@cmail.carleton.ca, yuhong.guo@carleton.ca}
 }

\begin{document}
\maketitle


\begin{abstract}
The generalization gap in reinforcement learning (RL) has been a significant 
obstacle that prevents the RL agent from learning general skills and adapting to varying environments. 
Increasing the generalization capacity of the RL systems can significantly improve their performance 
on real-world working environments. 
In this work, we propose a novel policy-aware adversarial data augmentation 
method to augment the standard policy learning method with automatically generated trajectory data. 
Different from the commonly used observation transformation based data augmentations,
our proposed method adversarially generates new trajectory data 
based on the policy gradient objective and 
aims to more effectively increase the RL agent's generalization ability 
with the policy-aware data augmentation. 
Moreover, we further deploy a mixup step to integrate the original and generated data
to enhance the generalization capacity while mitigating the over-deviation of the adversarial data. 
We conduct experiments 
on a number of RL tasks to investigate the generalization performance
of the proposed method 
by comparing it with the standard baselines and the state-of-the-art mixreg approach. 
The results show our method 
can generalize well with limited training diversity, 
and achieve the state-of-the-art generalization test performance.
\end{abstract}



\section{Introduction}
Benefited from the power of deep neural networks,
deep reinforcement learning (RL)  
has recently demonstrated
incredible performance on many human-level challenging tasks. 
In addition to traditional board games like Go~\citep{silver2016mastering, silver2017mastering}, 
deep reinforcement learning agents has defeated professional human players in large scale video games like StarCraft~\citep{vinyals2017starcraft} and Dota 2~\citep{berner2019dota}. 
Meanwhile,  
deep RL systems also suffer from the vulnerabilities of deep neural networks such as overfitting and data memorization,
which often induce generalization gaps between the training and testing performances of the RL agents. 
The generalization gap prevents the RL agents from learning general skills 
in a simulation environment
to handle the real-world working environments 
~\citep{cobbe2020leveraging,tobin2017domain}. 
As a result, 
deep RL agents are reportedly performing poorly on unseen environments,
especially when the training environments lack of diversity
\citep{zhang2018study,zhang2018dissection, cobbe2020leveraging, song2019observational, cobbe2019quantifying}. 

Towards the goal of reducing the generalization gap, 
previous works have exploited the 
conventional data augmentation techniques such as 
cropping, translation, and rotation to augment the input observations and 
increase the diversity of the training data in RL
\citep{shorten2019survey, cobbe2019quantifying, lee2019network, laskin2020reinforcement}.
Some have also explored the selection of data augmentation techniques
\citep{raileanu2020automatic, jiang2021prioritized} and improving the
RL architectures~\citep{raileanu2021decoupling, ball2021augmented}. 
Recently, a mixture regularization method has been introduced to 
learn generalizable RL systems~\citep{wang2020improving},
which deploys mixup to increase the data diversity and yields the state-of-the-art generalization performance. 
However, these methods work specifically at the level of the input observation data
without taking the RL system into consideration,
which prevents them from generating data that are most particularly suitable 
for the target RL system and hence could limit their performance's improvements. 

In this paper, we propose a novel policy-aware adversarial data augmentation method with Mixup enhancement (PAADA+Mixup)
to improve the generalization ability of policy learning based RL systems. 
This method first generates adversarial augmenting trajectory data by minimizing the expected rewards 
of the given RL policy based on the initial observed source trajectory data. 
Next it combines each augmenting adversarial trajectory and the corresponding observation source trajectory
together by randomly selecting each observation step from one of them. 
The combined trajectory data can then be used to update the policy of the RL system. 
As deep RL systems typically inherit the vulnerability of deep neural networks to adversarial examples \citep{lin2017tactics},  
some previous works have investigated the topic  
of adversarial attacks in deep RL~\citep{zhao2020blackbox, gleave2019adversarial}. 
Our work however is the first that deploys adversarial data augmentation in RL systems 
to improve their generalization capacity.  
Moreover, we further deploy a mixup operation on the combined trajectory 
to enhance the robustness and generalization of the RL system. 
We conduct extensive experiments on several RL tasks to investigate the proposed approach
under different generalization test settings. 
The experimental results show the proposed PAADA greatly outperforms the strong RL baseline method, 
proximal policy optimization (PPO)~\citep{schulman2017proximal}.  
With the mixup enhancement, PAADA+Mixup can 
achieve the state-of-the-art performance,
surpassing mixreg~\citep{wang2020improving} 
with notable performance gains.

The main contributions of the proposed work can be summarized as follows:
\begin{itemize}
\item 
We introduce adversarial data augmentation to deep RL
and develop the very first policy-aware adversarial data generation method to improve the generalization capacity of deep RL agents.
\item 
We integrate the generalization strengths of both adversarial data generation and mixup
and demonstrate superior performance than using either one of them alone. 
\item 
We conduct experiments on the Procgen benchmark with different generalization settings. 
Our proposed method demonstrates good generalization performance with limited 
training environments and outperforms the state-of-the-art mixreg approach. 
\end{itemize}


\section{Related Works}
\label{sec:related_works}

Generalization gap has been an increasing concern in deep reinforcement learning. 
Recent studies~\citep{zhang2018study, zhang2018dissection} show that the main cause of generalization gap is the over-fitting 
and memorization inherited from deep neural networks~\citep{arpit2017closer}. 
Data augmentation is a conventional technique for solving over-fitting in deep learning~\citep{shorten2019survey}. 
In recent studies, some broadly applied data augmentation approaches in deep learning have been brought into reinforcement learning. 
Cobbe et al.~\citep{cobbe2019quantifying} introduce the cutout technique into deep RL, 
where data are generated by partially blocking the input observations with randomly generated black occlusions. 
Lee et al.~\citep{lee2019network} propose a randomized convolutional network to perturb the input observations. 
Laskin et al.~\citep{laskin2020reinforcement} 
propose to integrate
the commonly used data augmentation skills such as cropping, translation, and rotation 
to improve the generalization of RL. 
Ball et al.~\citep{ball2021augmented} introduce augmented world models to 
specifically address the generalization of model-based offline RL problems. 
Raileanu et al.~\citep{raileanu2020automatic} propose a data-regularized actor-critic 
approach to regularize policy and value functions when applying data augmentation,
whereas the upper confidence bound~\citep{auer2002using} 
is borrowed to select the proper data augmentation method. 
In~\citep{wang2020improving}, 
Wang et al.
propose a simple but efficient mixture regularization (mixreg) approach to improve the generalization capacity in RL systems. 
Following the mixup~\citep{zhang2017mixup} in supervised learning, 
the mixreg method 
generates new observations from two randomly selected observations through linear combination. 
The deep RL agents trained on mixreg augmented observations demonstrate significant improvements on their generalizability.

Inspired by the success of data augmentation on generalization of deep reinforcement learning, 
we introduce a novel adversarial data augmentation technique to deep RL. 
Deep neural networks are known vulnerable to adversarial examples~\citep{szegedy2013intriguing}. 
Many previous works 
have studied the attack and defence strategies of 
deep neural networks to adversarial examples~\citep{yuan2019adversarial, madry2017towards, akhtar2018threat}. 
Lin et al. \citep{lin2017tactics} show deep reinforcement learning agents 
inherited the vulnerability of 
deep neural networks to adversarial examples. 
They propose two tactics to perform adversarial strategies to attack RL agents: 
the strategically-timed attack minimizes the agent's reward for small subset of time steps, 
while the enchanting attack lures the agent to a specific target state. 
Following this, a few works have further contributed to this rising topic of adversarial attacks in deep RL. 
Zhao et al.~\citep{zhao2020blackbox} propose an approach to generate adversarial observations without previous knowledge on the network architecture and RL algorithm of the deep reinforcement learning agent. Instead of adding perturbations to the observations, Gleave et al.~\citep{gleave2019adversarial} propose to train the agent with an adversarial policy.

Distinct from adversarial attacks, our study focuses on adversarial data augmentation.
Adversarial data augmentation has been investigated in a number of works on supervised learning,
but has not been explored to enhance generalization in deep RL systems. 
Goodfellow et al.~\citep{goodfellow2014explaining} find the linear property of the deep neural network is vulnerable to 
adversarial perturbations 
and propose to train the supervised model with adversarial examples to improve generalization. 
Sinha et al.~\citep{sinha2017certifying} propose a Lagrangian formulation of adversarial perturbations in a Wasserstein ball~\citep{lee2017minimax} to enhance the robustness of deep learning models. 
Volpi et al.~\citep{volpi2018generalizing} further propose to exploit adversarial data augmentation 
for domain adaptation with unknown target domains. 


\section{Method}
\label{sec:method}

This study focuses on increasing the generalization ability of a deep RL agent. 
Following the previous generalization study in RL~\cite{wang2020improving}, 
we consider the following RL setting. 
The agent is trained on a set of $n$ environments $\{\mathcal{K}_1,...,\mathcal{K}_n\}$ 
sampled from a distribution $p(\mathcal{K})$ to learn an optimal policy $\pi^*$,
and then tested on another set of 
environments $\{\hat{\mathcal{K}}_1,...,\hat{\mathcal{K}}_m\}$ sampled from $p(\mathcal{K})$. 
Its generalization performance is measured as the zero-shot expected cumulative reward in the test environments:
\begin{equation}
\label{eq:gen}
	\mathbb{E}_{\tau \sim \mathcal{D}_{{\pi}^*}^{test}}\sum_{t=0}^T\gamma_*^{t} r_t
\end{equation}
where $\tau$ denotes a trajectory $(s_0, a_0, r_0, s_1, a_1, r_1,\cdots, r_T)$,
$\mathcal{D}_{{\pi}^*}^{test}$ denotes the distribution of $\tau$ in the test environments under policy $\pi^*$,
and $\gamma_*\in(0,1]$ is the discount factor. 
Moreover, we assume the training environments can be much fewer than the test environments 
such as $n =\lceil \xi m\rceil$ with $\xi\in (0,1]$. 
The goal is to train an optimal policy $\pi^*$ that can generalize well in terms of the expected cumulative test reward above in Eq.(\ref{eq:gen}).
Towards this goal, in this section we present a policy-aware adversarial data augmentation with mixup enhancement (PAADA+Mixup) method
for the RL training process. 

\subsection{Policy-Aware Adversarial Data Augmentation}
When the training and test environments are different, 
a standard RL learning algorithm, e.g,
Proximal Policy Optimization (PPO)~\citep{schulman2017proximal},
will inevitably suffer from the domain gap between the training and test environments,
and demonstrate generalization gaps between the training and test performances.
Inspired by the effectiveness of adversarial data augmentation in supervised learning~\citep{volpi2018generalizing}, 
we propose to augment the deep RL process by generating adversarial trajectories from the current policy,
aiming to adaptively broaden the experience of the RL agent and increase its generalization capacity 
to unseen test environments.

Specifically, given the current parametric policy $\pi_\theta$ with parameter $\theta$, 
in each epoch of policy optimization based RL training, 
the standard procedure is to collect a set of trajectories $\{\tau_1,\cdots, \tau_n\}$ from the training environments,
and then update the policy parameter $\theta$ by performing gradient ascent
with respect to the policy optimization objective
over the observed trajectories. 
For policy gradient, the following surrogate objective is often used:
\begin{align}
	L^{PG}(\theta) = \mathbb{\hat{E}}_t [\log\pi_\theta(a_t|s_t)\hat{A}_t]
\end{align}
where $\mathbb{\hat{E}}_t[\cdot]$ denotes the empirical average over the set of transitions in the collected trajectories; 
$\hat{A}_t$ is the estimated advantage function 
at timestep $t$,
and can be approximated as $\hat{A}_t=r_t-V(s_t)$, where $V(s_t)$ is the value function at state $s_t$
\cite{degris2012model}.  
For the more advanced high-performance policy gradient algorithm PPO~\citep{schulman2017proximal}, 
a clipping modulated objective will be typically used:
\begin{align}
	\!\!
	L^{\mbox{\small\it PPO-C}}(\theta) 
	&\!=\! \mathbb{\hat{E}}_t\!\left[\min\!\left(\rho_\theta\hat{A}_t,
	\mbox{clip}\left(\rho_\theta, 1\!-\!\epsilon, 1\!+\!\epsilon \right)\!\hat{A}_t
	\right)\right]
	\label{eq:PPOclip}
\end{align}
where $\rho_\theta=\frac{\pi_\theta(a_t|s_t)}{\pi_{\theta'}(a_t|s_t)}$
and $\epsilon$ is a small constant.
This PPO objective regularizes the new policy $\pi_\theta$ from being severely deviated from
the previous policy $\pi_{\theta'}$ and aims to avoid large destructive policy updates 
associated with the vanilla policy gradient.  
Although PPO has demonstrated great performance in standard RL, it may be overly bounded
to the available training observations and hence yields poor test performance 
in the generalization settings. 
We therefore propose to expand the observation space by generating adversarial examples
based on the collected trajectories. 

\subsubsection{Adversarial Trajectory Generation}
For each observed trajectory $\tau$, 
we generate an adversarial example 
for each of its observation points (i.e., transitions),  
$P_t = (x_t, y_t)$ with $x_t = s_t$ and 
$y_t = (a_t, r_t)$.
That is, we find the worst example $P=(x, y)$ in the close neighborhood of the current
point $P_t= (x_t, y_t)$ by minimizing the cumulative award objective $L$ the RL agent needs to maximize, such as
\begin{align}
	\min_{P}\; L(\theta; P)\qquad \mbox{s.t.}\; D(P, P_t)\leq\rho
	\label{eq:AdvCon}
\end{align}
where $D(\cdot,\cdot)$ denotes a distance metric such as the Wasserstein distance
and the constraint bounds the point $P$ to be within the neighborhood of $P_t$. 
For two points $P=(x,y)$ and $P_t=(x_t,y_t)$, the Wasserstein distance $D(P, P_t)$ can be specified as
follows through a transportation cost $c$
\cite{volpi2018generalizing}:
\begin{align}
D(P,P_t)
=&\, c((x,y), (x_t,y_t))
=
||x-x_t||^2+\infty\cdot\mathbf{1}\{y\neq y_t\}
\label{equation:adv3}
\end{align}
Moreover, the constrained problem in Eq.(\ref{eq:AdvCon}) can be equivalently reformulated
as the following regularized optimization problem with a proper Lagrangian parameter $\gamma$: 
\begin{align}
	&\min_P\; L(\theta;P) + \gamma D(P,P_t) 
	\quad \iff \quad \min_x L(\theta;x,y_t) + \gamma \|x-x_t\|^2
\label{equation:adv2}
\end{align}
As this point generation is conducted under the current policy $\pi_\theta$ and does not involve policy update,
we use the policy gradient objective $L^{PG}$ as the objective $L$ for adversarial example generation
due to its simplicity and easy computation.  
Therefore for each observation point $(s_t, a_t, r_t)$ in the collected trajectory data, 
its corresponding adversarial point $(\hat{s}_t, a_t, r_t)$ will be generated as follows:
\begin{align}
\hat{s}_t
&=\underset{s}{\mathrm{argmin}}~L^{PG}(\theta;s,t)+\gamma D(s,s_t)\nonumber\\
&=\underset{s}{\mathrm{argmin}}~\log\pi_\theta(a_t|s)\hat{A}_t +\gamma \|s-s_t\|^2\nonumber\\
&=\underset{s}{\mathrm{argmin}}~\log\pi_\theta(a_t|s)(r_t-V(s))+\gamma \|s-s_t\|^2
\label{eq:advS}
\end{align}
We can solve this generation problem by performing gradient descent with a stepsize $\eta$ and a maximum step number $K_{max}$. 
The algorithm is shown in Algorithm~\ref{alg:generate}.
\begin{algorithm}[t]
\caption{Adversarial Observation Generation}
\label{alg:generate}
\hspace*{\algorithmicindent}
\textbf{Input:}\; $\pi_\theta, V$,\, $(s_t,a_t, r_t)$;\; stepsize $\eta$,\;
	maximum step number $K_{max}$,\; tolerance $\epsilon_a$\\ 
\hspace*{\algorithmicindent}
\textbf{Output:} adversarial observation $s$
\begin{algorithmic}[1]
\State $s =s_t$	
\For{$k=1,...,K_{max}$}
	\State{\bf if}  {$\|\nabla_s[\log\pi_\theta(a_t|s)(r_t-V(s)) +\gamma (s-s_t)^2]\|^2<\epsilon_a$} 
	\hspace*{1.6em}{\bf then} break
\State $s=s-\eta\nabla_s[\log\pi_\theta(a_t|s)(r_t-V(s)) +\gamma (s-s_t)^2]$
\EndFor
\end{algorithmic}
\end{algorithm}
With this simple procedure,
for each observation trajectory $\tau$, 
we can produce a corresponding adversarial trajectory $\hat{\tau}$
by generating an adversarial point $(\hat{s}_t, \hat{a}_t=a_t, \hat{r}_t=r_t)$ 
for each observation point $(s_t, a_t, r_t)$ in $\tau$.  

\subsubsection{Trajectory Augmentation}
Given the observed source trajectory $\tau$ and the generated adversarial trajectory $\hat{\tau}$,
simply appending one after the other to fed to the deep RL agent for training turns out not to be a suitable solution,
as it may cause the policy parameter update to dramatically switch between very different directions. 
Moreover, the degree of augmentation could also matter for different RL tasks. 
Here we propose to augment the source trajectory $\tau$ with the adversarial trajectory $\hat{\tau}$
by combining them into a new trajectory $\overline{\tau}$ with an augmentation degree $\nu\in[0,1]$.
Specifically, we construct the new trajectory $\overline{\tau}$ by randomly selecting 
$\lfloor\nu|\tau|\rfloor$ points (transitions) from the adversarial trajectory $\hat{\tau}$ 
and taking the other $\lceil(1- \nu)|\tau|\rceil$ points from the original trajectory $\tau$.
In this way, we not only can better blend the adversarial points with the original observations, 
but also have control over the contribution degree of the augmentation data through the hyperparameter $\nu$.

We deploy this policy-aware adversarial data augmentation scheme on the PPO method. 
The overall training algorithm is depicted in Algorithm~\ref{algorithm:procedure},
which first pre-trains the RL agent for $K_{pre}$ epochs 
and then performs adversarial data augmentation in ensuing epochs. 

\begin{algorithm}[t]
\caption{Adversarial Data Augmentation on PPO}
\label{algorithm:procedure}
\hspace*{\algorithmicindent}
\textbf{Input:} initial policy parameter $\theta$,\; initial value function 
	parameter $\phi$,\, \\
\hspace*{\algorithmicindent}
	\hspace{2.6em} 
	the pre-training epoch number $K_{pre}$,\; 
	the augmentation degree $\nu$
\\
\hspace*{\algorithmicindent}
\textbf{Output:} trained policy $\pi_\theta$
\begin{algorithmic}[1]
\For{$k=1,2,...$}
\State Collect a set of trajectories $\{\tau_1,\tau_2,...,\tau_n\}$ by
	running policy $\pi_\theta$ on the training environments %
\For{$\tau_i$ in $\{\tau_1,\tau_2,...,\tau_n\}$}
	\State Compute the advantage estimates $\{A_t\}$ for all 
	the $t$ transition points in $\tau_i$
\If{$k < K_{pre}$}
\State continue
\EndIf
\For{$t=0,1,2,...,|\tau_i|-1$}
	\State Generate the adversarial state $\hat{s}_t$ with Eq.(\ref{eq:advS})
	\State $\hat{A_t}=r_t-V_\phi(\hat{s}_t)$
	\State Add $(\hat{s_t},a_t,r_t,\hat{A_t})$ into the augmentation 
	trajectory $\hat{\tau}_i$
\EndFor
	\State Combine $\tau_i$ and $\hat{\tau}_i$ into an augmented trajectory $\overline{\tau}_i$ with augmentation degree $\nu$
	\State \%[place holder for additional step]
\EndFor
\State Update the policy function parameter $\theta$ by maximizing 
	the PPO-Clip objective in Eq.(\ref{eq:PPOclip}) on the augmented trajectories $\{\bar{\tau}_1,\bar{\tau}_2,...,\bar{\tau}_n\}$
\State Update the value function parameter $\phi$ on the augmented trajectories
\EndFor
\end{algorithmic}
\end{algorithm}
\subsection{Enhancement with Mixup}
In addition to the adversarial data augmentation technique above, 
we consider further enhancing the diversity of the training data with a Mixup procedure. 
Mixup generates data points through linear interpolation and has demonstrated effective performance in 
both supervised learning~\citep{zhang2017mixup} and reinforcement learning~\citep{wang2020improving}. 
Here we propose to deploy the Mixup procedure on each augmented trajectory $\overline{\tau}_i$ generated above. 
Specifically, for a trajectory $\overline{\tau}_i$ with $|\overline{\tau}_i|$ transition points, 
we first make a copy of $\overline{\tau}_i$ as $\overline{\tau}'_i$, and  
randomly shuffle the indices $\{0,1,\cdots,|\overline{\tau}_i|\}$ into $\{I_0,I_1,\cdots,I_{|\overline{\tau}_i|}\}$.
Then we linearly combine $\overline{\tau}_i$ and $\overline{\tau}'_i$ with the following mixup steps:
\begin{align}
 \bar{s}_t=\lambda\bar{s}_t+(1-\lambda)\bar{s}'_{I_t}\\
 \bar{r}_t=\lambda\bar{r}_t+(1-\lambda)\bar{r}'_{I_t}\\
 \bar{A}_t=\lambda\bar{A}_t+(1-\lambda)\bar{A}'_{I_t}
\end{align}
while $\bar{a}_t$ is set as $\bar{a}_t$
 with probability $\lambda$ and set as $\bar{a}'_{I_t}$ with probability $1 - \lambda$. 
The hyperparameter $\lambda$ is sampled from a beta distribution $\lambda\sim B(\alpha, \beta)$.
Normally, the parameters of the beta distribution are set to $\alpha=\beta$ as suggested in~\citep{zhang2017mixup}. 
However, when the training environments are limited, 
it is beneficial to have different $\alpha$ and $\beta$ values to 
shift the mean value of the $\lambda$ samples. 
This Mixup step can be deployed on the augmented trajectory
and performed on line 12 
 within the trajectory loop in Algorithm~\ref{algorithm:procedure}.


\section{Experiments}
\label{sec:experiments}
We conducted experiments to validate the empirical performance
of the proposed method under different generalization settings.
In this section, we report our experimental settings and results. 

\subsection{Experiment Setting}
We conduct experiments
on the Procgen benchmark~\citep{cobbe2020leveraging}, which 
contains procedurally generated environments designed to test the generalization ability of deep RL agents. 
Each environment takes visual input and has significant change among different levels of the environments. 
A Procgen environment 
can generate a maximum of 500 different levels for the RL generalization task. 
We choose 4 game environments (starpilot, dodgeball, climber, fruitbot) from this benchmark as different RL tasks
and treat different levels of each RL task as different training and test environments from the generalization perspective.
Follow the settings in~\citep{cobbe2020leveraging,wang2020improving}, 
we do not limit the levels in the testing environments and use the total $m=500$ levels for testing,
while a relatively smaller number, $n=\lceil \xi m\rceil$, of level environments, are sampled for training. 
In particular, we consider different $\xi$ values such as $\xi \in \{0.25, 0.5, 1\}$.
A smaller $\xi$ value indicates a more difficult generalization setting as the diversity of training environments
is reduced. 

We adopt PPO~\cite{schulman2017proximal} as our RL baseline,
although the proposed methodology in principle can be generalized into other RL methods as well.
In addition, we compare our approach, PAADA+Mixup, to the 
state-of-the-art generalization method, mixreg, 
which has been shown to outperform the conventional data augmentation techniques in~\citep{wang2020improving}.
Following the Procgen benchmark, we adopt the same convolutional neural network architecture as IMPALA\citep{espeholt2018impala}. 
We use the mean episode return in each epoch of the zero-shot testing as the generalization evaluation metric on each of the four Procgen benchmark games
(starpilot, dodgeball, climber, fruitbot). 
Moreover, following~\citep{cobbe2020leveraging}
we compute the mean normalized return over the four games to summarize the overall generalization performance. 
%

\begin{figure*}[t]
\centering
\includegraphics[width=1\textwidth]{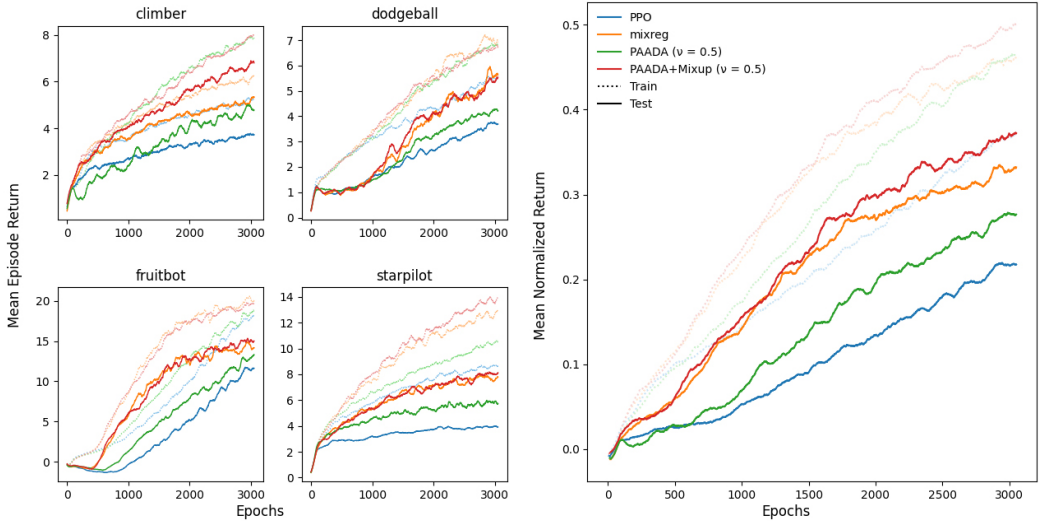}
\caption{Training and testing performance on full training environments ($\xi=1$). 
\textbf{Left}: Mean episode return per epoch on different game environments. 
\textbf{Right}: Mean normalized return over all the four game environments. }
\label{fig:nenv100}
\end{figure*}

\paragraph{Hyperparameters}
For the adversarial observation generation, the stepsize $\eta$ is set to 10,
the maximum number of steps $K_{max}$ is set to 50, 
and the tolerance $\epsilon_a$ is set to $5e^{-6}$.
For the training algorithm of the proposed approach, PAADA, 
we set the pre-training epoch number, $K_{pre}$, as 50. 
That is, in the first $50$ training epochs, the deep RL agent is trained on standard PPO~\citep{schulman2017proximal}. 
After 50 epochs, our augmentation method is applied to enhance the generalizability. 
In the evaluation plots, the epoch number does take these pre-training epochs into account. 
The Lagrangian hyperparameter $\gamma$ in Eq.(\ref{equation:adv2}) and Eq.(\ref{eq:advS}) is set to $0.01$. 
In each training epoch, we collect $n$ trajectories and each trajectory has $256$ transitions. %
For testing, $m$ trajectories, one from each testing environment, are used.
The discount factor $\gamma_*$ is set to 0.999. 
Each experiment is tested on 3054 training epochs 
in total. 
For the Mixup procedure, we use $\alpha=0.2$ and $\beta=0.2$ for $\xi=1$, 
increase $\beta$ to $\beta=0.5$ for $\xi=0.5$ and to $\beta=1$ for $\xi=0.25$.


\begin{figure*}[t]
\centering
\includegraphics[width=1\textwidth]{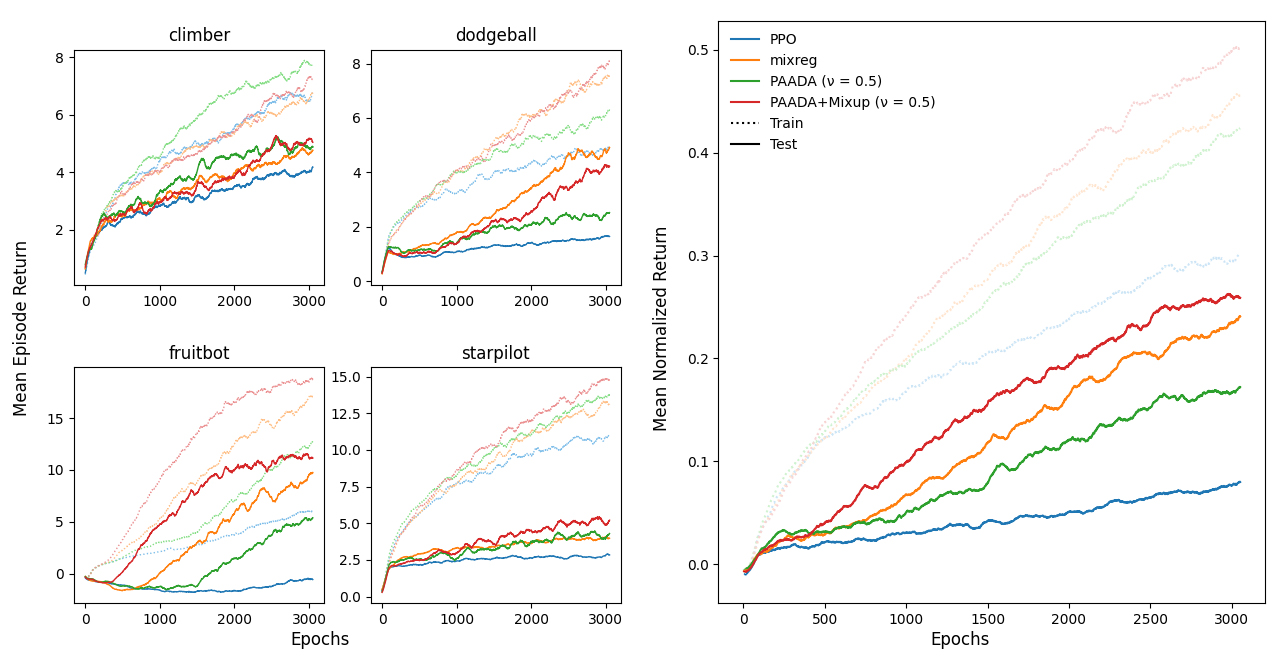}
\caption{Training and testing performance on partial training environments ($\xi=0.5$). 
\textbf{Left}: Mean episode return per epoch on different game environments. 
\textbf{Right}: Mean normalized return over all the four game environments}
\label{fig:nenv50}
\end{figure*}

\begin{figure*}[t]
\centering
\includegraphics[width=1\textwidth]{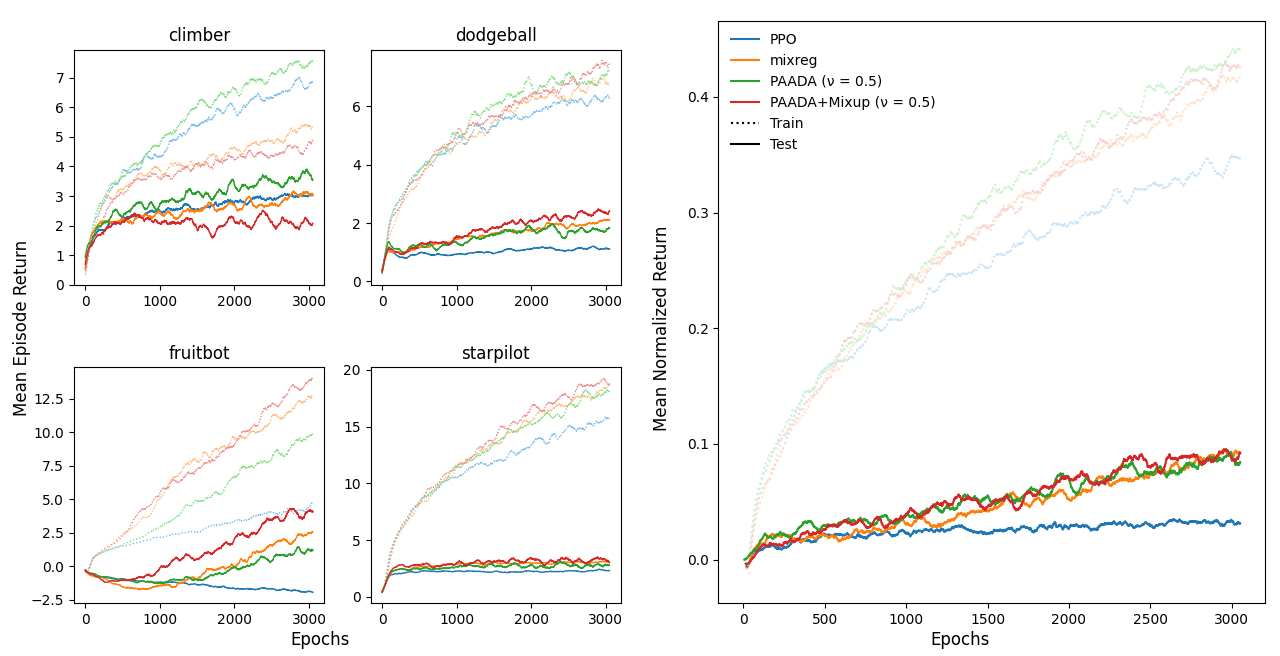}
\caption{Testing performance on partial training environments ($\xi=0.25$). 
\textbf{Left}: Mean episode return per epoch on different game environments. 
\textbf{Right}: Mean normalized return over all the four game environments}
\label{fig:nenv25}
\end{figure*}

\subsection{Experiments with Full Set of Training Environments}

In the first set of less challenging experiments, for each game, we have $\xi=1$ and 
use the full set of 500 training environments,
while testing on unlimited environments. 
For the proposed approach PAADA, we tested two different variants of it:
(1) PAADA ($\nu=0.5$), which uses an augmentation degree $\nu=0.5$ and indicates that 
50\% of the observation transitions of the adversarial trajectory are randomly merged into 
the original observation trajectory. 
(2) PAADA+Mixup ($\nu=0.5$), which indicates the Mixup enhancement is added above PAADA ($\nu=0.5$).
We compared these variants with the PPO baseline and mixreg.

The training and test evaluation results on the four games are reported in Figure~\ref{fig:nenv100}.
The left four plots report the comparison results on the four games separately. 
We can see that 
the policy-aware adversarial data augmentation alone outperforms the baseline PPO -- 
PAADA$(\nu=0.5)$ greatly improve the generalization performance of PPO on the four games
and the performance gain in terms of the mean normalized return is very notable.
With the Mixup enhancement, PAADA+Mixup$(\nu=0.5)$ further improves its generalization performance,
and it yield similar performance to 
the state-of-the-art method, mixreg, on three games, {\em dodgeball, fruitbot} and {\em starpilot},
and outperforms mixreg on {\em climber} with large margins.
The right plot in Figure~\ref{fig:nenv100} summarizes the overall performance over the four games in terms of a mean normalized return metric. 
It shows PAADA+Mixup$(\nu=0.5)$ outperforms 
all the other methods, including mixreg, with distinct performance gains. 
These results clearly demonstrate the efficacy of the proposed adversarial based data augmentation and mixup.

\subsection{Experiments on Partial Set of Training Environments}
In this set of experiments, we consider more challenging generalization settings with $\xi < 1$.
That is, the number of training environments are much smaller and hence the training diversity is greatly reduced. 
In particular, we consider $\xi = 0.5$ and $\xi=0.25$. 
In such settings, 
with limited training diversity, it is more difficult to achieve better generalization. 

Again we compared the two variants of the proposed method, PAADA$(\nu=0.5)$ and PAADA+Mixup$(\nu=0.5)$, 
with PPO and mixreg on the four game tasks. 
Figure~\ref{fig:nenv50} reports the testing results of the comparison methods with $50\%$ training environments, i.e., $\xi=0.5$. 
We can see that in this setting, 
both PAADA$(\nu=0.5)$ and PAADA+Mixup$(\nu=0.5)$ outperform PPO on 
all the four games with clear performance gains. 
Between the two variants, the mixup still bring certain improvements on three games, 
{\em starpilot, dodgeball} and {\em fruitbot},
whereon PAADA+Mixup$(\nu=0.5)$ outperforms PAADA$(\nu=0.5)$.
Moreover, 
PAADA+Mixup$(\nu=0.5)$ outperforms mixreg on three games, 
{\em climber, fruitbot} and {\em starpilot},
while mixreg produces the best performance on {\em dodgeball}.
Nevertheless, as shown 
in the right plot of Figure~\ref{fig:nenv50}, 
PAADA+Mixup$(\nu=0.5)$ demonstrates a consistent and clear advantage over mixreg in terms of the overall mean normalized return. 

Figure~\ref{fig:nenv25} reports the results of the comparison methods with $25\%$ training environments, i.e., $\xi=0.25$. 
This setting is more challenging than the above one with $\xi=0.5$. 
In this setting, we can see that 
PAADA+Mixup$(\nu=0.5)$ 
still outperforms mixreg on {\em dodgeball, fruitbot} and {\em starpilot},
but produces inferior result on {\em climber},
whereon PAADA$(\nu=0.5)$ produces the best results. 
In terms of mean normalized return, as shown in the right plot of Figure~\ref{fig:nenv25},
all three data augmentation methods demonstrate similar superior generalization capacity over the baseline PPO,
All these experiments show our propose adversarial data augmentation with mixup enhancement can effectively 
increase the generalization capacity of RL systems.


\begin{table*}[t!]
\vskip .2in	
\centering
	\resizebox{\textwidth}{!}{
\begin{tabular}{lccccc}
\Xhline{1pt}
Environment ($\xi=1$)	& PPO   & mixreg   & PAADA ($\nu=0.5$)   & PAADA+Mixup ($\nu=0.5$) \\ \hline
climber & $3.74\pm 0.56$ & $5.31\pm 0.57$ & $4.79\pm 0.55$ & $\mathbf{6.79\pm 0.59}$ \\
dodgeball & $3.70\pm 0.45$ & $\mathbf{5.64\pm 0.62}$ & $4.24\pm 0.42$ & $5.48\pm 0.56$ \\
fruitbot & $11.58\pm 1.34$ & $14.12\pm 1.40$ & $13.19\pm 1.12$ & $\mathbf{14.89\pm 1.18}$ \\
starpilot & $3.94\pm 0.46$ & $7.67\pm 0.84$  & $5.74\pm 0.92$ & $\mathbf{8.09\pm 0.87}$ \\
MNR & $0.22\pm 0.02$ & $0.33\pm 0.02$ & $0.28\pm 0.02$ & $\mathbf{0.37\pm 0.02}$ \\
\Xhline{1pt}
Environment ($\xi=0.5$)	& PPO   & mixreg    & PAADA ($\nu=0.5$)   & PAADA+Mixup ($\nu=0.5$) \\ \hline
climber & $4.14\pm 0.52$ & $4.72\pm 0.51$ & $4.87\pm 0.54$ & $\mathbf{5.09\pm 0.56}$ \\
dodgeball & $1.64\pm 0.23$ & $\mathbf{4.93\pm 0.54}$  & $2.53\pm 0.44$ & $4.23\pm 0.67$ \\
fruitbot & $-0.55\pm 0.58$ & $9.66\pm 1.19$ &  $5.26\pm 1.27$ & $\mathbf{11.13\pm 1.32}$ \\
starpilot & $2.85\pm 0.34$ & $3.98\pm 0.39$ & $4.27\pm 0.88$ & $\mathbf{5.10\pm 0.94}$ \\
MNR & $0.08\pm 0.01$ & $0.24\pm 0.02$ & $0.17\pm 0.02$ & $\mathbf{0.26\pm 0.02}$ \\
\Xhline{1pt}
Environment ($\xi=0.25$)	& PPO   & mixreg    & PAADA ($\nu=0.5$)   & PAADA+Mixup ($\nu=0.5$) \\ \hline
climber & $3.05\pm 0.47$ & $3.07\pm 0.55$ & $\mathbf{3.64\pm 0.62}$ & $2.08\pm 0.46$ \\
dodgeball & $1.12\pm 0.18$ & $2.10\pm 0.30$ & $1.82\pm 0.47$ & $\mathbf{2.38\pm 0.46}$ \\
fruitbot & $-1.90\pm 0.39$ & $2.48\pm 0.81$ & $1.19\pm 0.98$ & $\mathbf{4.02\pm 1.06}$ \\ 
starpilot & $2.30\pm 0.28$ & $3.03\pm 0.34$ & $2.79\pm 0.91$ & $\mathbf{3.13\pm 1.02}$ \\
MNR & $0.03\pm 0.01$ & $\mathbf{0.09\pm 0.01}$ & $\mathbf{0.09\pm 0.02}$ & $\mathbf{0.09\pm 0.02}$ \\
\Xhline{1pt}
\end{tabular}%
}
\caption{\label{tab:datasets}Mean and standard deviation of the average test returns among 100 test epochs. 
	MNR: mean normalized return over all the four games, 
	which is a criterion shows the overall performance of a generalization method. 
	The best result in each setting is shown in bold font.
	}
\label{table:paada}
\vskip .1in	
\end{table*}

\subsection{Overall Generalization Capacity Comparison}

To demonstrate a more concrete comparison between all the methods and across different settings and games, 
we collected the test results for the last 100 of the 3054 training epochs
and reported the mean and standard deviation of these test returns in Table~\ref{table:paada}.
From these concrete result numbers 
we can see that all the three methods with generalization strategies, 
PAADA$(\nu=0.5)$, PAADA+Mixup$(\nu=0.5)$, and mixreg, consistently 
outperform PPO across different game tasks and different generalization settings. 
This suggests both adversarial data augmentation and mixup are individually effective
in improving the generalization performance of the baseline RL. 
By effectively integrating both adversarial data augmentation and mixup enhancement,
our proposed PAADA+Mixup$(\nu=0.5)$ outperforms the state-of-the-art mixreg across almost all cases 
(12 out of 15 cases)
except on dodgeball with $\xi=1, 0.5$ and on climber with $xi=0.25$.
These results again validated the efficacy of the proposed method.

\section{Conclusion}
In this paper, we proposed a novel policy-aware adversarial data augmentation method with
Mixup enhancement (PAADA+Mixup) to improve the generalization capacity of RL systems. 
It generates augmenting trajectories by adversarially minimizing the expected reward that 
a RL agent would desire to maximize,
and then uses them to augment the original trajectories under controlled augmentation degree. 
Moreover, a mixup operation is further deployed to enhance the diversity of the augmented trajectory. 
This is the first work that deploys adversarial data augmentation to learn generalizable RL systems.
It also presents the first experience of integrating adversarial augmentation and mixup generalization. 
We conducted experiments on the Procgen benchmark by comparing the proposed method with both 
the baseline PPO 
and the state-of-the-art method 
under different generalization settings. 
The results show PAADA surpasses PPO in general, while
PAADA+Mixup outperforms the state-of-the-art mixreg with notable performance gains,
especially in the most challenging generalization setting. 
%

\bibliography{paperbib}
\end{document}